\documentclass[runningheads]{llncs}

\usepackage[T1]{fontenc}
\usepackage{graphicx}
\usepackage{amsmath}
\usepackage{amssymb}
\usepackage{booktabs}
\usepackage{algorithm}
\usepackage{algorithmic}
\usepackage{listings}
\usepackage{xcolor}
\usepackage{multirow}
\usepackage{adjustbox}
\usepackage{placeins}
\usepackage{subcaption}
\usepackage[hidelinks]{hyperref}
\usepackage{url}

\lstdefinestyle{prolog}{
  basicstyle=\ttfamily\small,
  keywordstyle=\bfseries,
  commentstyle=\color{gray},
  frame=single,
  breaklines=true,
  xleftmargin=1em,
  xrightmargin=1em,
  aboveskip=6pt,
  belowskip=6pt
}

\title{ALBI: LLM-Guided Inductive Logic Programming via Iterative Generate-Test-Constrain Refinement}

\titlerunning{ALBI: LLM-Guided ILP}

\author{Aboubaker Seddiq Benterki\inst{1}  \and
J\"uri Vain\inst{1}\and
Gert Kanter\inst{1} 
 \and
Leelo Aino J\"urgenstein\inst{1}}

\authorrunning{A.\,S. Benterki et al.}

\institute{Tallinn University of Technology (TalTech), Ehitajate tee 5, 19086 Tallinn, Estonia\\
\email{\{aboubaker.benterki, leelo.jurgenstein, gert.kanter, juri.vain\}@taltech.ee}}

\begin{document}
\maketitle

\begin{abstract}
We introduce ALBI (Agent for Learning By Induction), a neuro-symbolic ILP framework that replaces the ASP-based hypothesis generator in the Learning from Failures (LFF) paradigm with a Large Language Model (LLM), while retaining SWI-Prolog as a formal verifier.
ALBI implements an iterative \emph{generate-test-constrain} loop: the LLM proposes candidate hypotheses from structured prompts; Prolog tests each candidate; and diagnostic feedback drives refinement.
We further introduce \emph{semantic anonymization}, a metamorphic testing methodology, to distinguish structural symbolic induction from surface-level pattern matching on predicate names.
Experiments on 30 standard ILP benchmarks show that ALBI achieves \textbf{100\%} success across three frontier LLMs (Gemini~3.1~Pro, GPT~5.4~Pro, Claude~Opus~4.6), matching Popper while requiring only 1.4--1.7 iterations on average.
Under anonymization, all models maintain 100\% success with a moderate iteration increase (1.36--1.67 times), providing evidence of robust symbolic reasoning.

\end{abstract}

\section{Introduction}
\label{sec:intro}

Inductive Logic Programming (ILP) addresses the fundamental challenge of inducing first-order logic theories from examples and background knowledge~\cite{muggleton1994ilp}.
Unlike statistical learning methods that produce opaque models, ILP yields human-readable, verifiable logic programs making it particularly suited for domains requiring explainability and formal guarantees.
The Learning from Failures (LFF) paradigm~\cite{cropper2021learning}, implemented in the Popper system, decomposes ILP into three stages: \emph{generate} a candidate hypothesis, \emph{test} it against examples, and \emph{constrain} the search space based on failures.
This approach has achieved strong results across diverse domains, but its reliance on ASP solvers for hypothesis generation introduces a fundamental bottleneck: the combinatorial explosion of the search space as background knowledge complexity grows~\cite{cropper2022ilp30}.

Recent advances in Large Language Models (LLMs) have demonstrated remarkable capabilities in code generation and logical reasoning~\cite{wei2022chain}. Gandarela et al.~\cite{gandarela2025expressivity} provide the first systematic analysis of LLM capabilities in logic theory induction, revealing that while LLMs show competitive performance on simpler theories, they struggle with long predicate chains and formal consistency. These are precisely the weaknesses that a symbolic verifier can detect and correct, suggesting a natural division of labor: leveraging the LLM for \emph{hypothesis generation} and the formal logic engine for \emph{verification and constraint enforcement}.

We introduce \textbf{ALBI} (Agent for Learning By Induction), a framework that operationalizes this insight within the LFF paradigm.\footnote{Implementation and prompt templates: \url{https://github.com/seddiqbenterki/albi}}
ALBI replaces the ASP generator with an LLM, augmented by:
(i)~a structured prompt architecture incorporating background knowledge, examples, bias constraints, and accumulated failure feedback;
(ii)~beam search over multiple candidate hypotheses with MDL-based selection~\cite{rissanen1978mdl}, mitigating mode collapse in LLM generation~\cite{zhang2025verbalized};
(iii)~an incremental \emph{locked clause} mechanism for learning multi-clause theories;
and (iv)~a divide-and-conquer optimization phase guided by the Minimum Description Length principle.

A critical concern with LLM-based ILP is whether the model performs structural symbolic induction or merely exploits semantic cues in predicate names.
We address this through \emph{semantic anonymization}~\cite{chen1998metamorphic,chen2018metamorphic}, a metamorphic testing methodology that replaces all meaningful predicate names with abstract identifiers, enabling rigorous evaluation of the system's symbolic reasoning capability.

Our contributions are:
\begin{enumerate}
    \item A neuro-symbolic ILP framework (\textbf{ALBI}) that integrates LLMs into the LFF paradigm as heuristic hypothesis generators with formal SWI-Prolog verification.
    \item A beam search strategy with structured diagnostic feedback that iteratively constrains the LLM's generation.
    \item A semantic anonymization methodology for metamorphic testing of LLM-based ILP systems.
    \item Empirical evaluation on 30 benchmark tasks with semantic and anonymized variants, comparing three frontier LLMs against the ASP-based Popper system.
\end{enumerate}

\section{Related Work}
\label{sec:related}

\subsection{Inductive Logic Programming}

ILP~\cite{muggleton1994ilp} has evolved over three decades~\cite{cropper2022ilp30}, from inverse entailment~\cite{muggleton1995progol} and bottom-up generalization~\cite{srinivasan2001aleph} to meta-interpretive learning~\cite{muggleton2014metagol,muggleton2015mil} and answer set programming~\cite{law2014ilasp,law2020ilasp3}.
The LFF approach~\cite{cropper2021learning} (Popper) learns constraints from test failures; subsequent extensions add divide-and-conquer strategies~\cite{cropper2022dcc}, search pruning~\cite{cropper2023wherenot}, rule joining~\cite{korner2024bigjoining}, program combination~\cite{cropper2023combo}, higher-order abstractions~\cite{hocquette2024stevie}, and noisy-data handling~\cite{hocquette2024noisy}.
Our locked clause mechanism shares the spirit of program combination but uses the LLM as the combination engine.

\subsection{LLMs for Logical Reasoning and Neuro-Symbolic AI}

Chain-of-thought prompting~\cite{wei2022chain} and iterative self-refinement~\cite{madaan2023selfrefine} showed that LLMs can perform multi-step reasoning when guided by structured feedback.
Gandarela et al.~\cite{gandarela2025expressivity} evaluated LLMs on ILP tasks, finding that formal inference feedback can approach ILP system performance on simpler theories but degrades on complex rule dependencies.
Our work embeds the LLM within a formal generate-test-constrain loop, providing the symbolic grounding that purely neural approaches lack.
More broadly, the integration of neural and symbolic AI is a key direction for robust AI~\cite{garcez2023neurosymbolic,hamilton2024nesysurvey}. Differentiable ILP ($\partial$ILP~\cite{evans2018dilp}) and neural-abductive approaches~\cite{dai2019abductive} learn logic programs via gradient methods but sacrifice search interpretability.
ALBI falls within the \emph{neural generator + symbolic verifier} paradigm, preserving soundness guarantees while leveraging neural semantic heuristics.

\subsection{Metamorphic Testing}

Metamorphic testing~\cite{chen1998metamorphic,chen2018metamorphic} addresses the oracle problem by defining properties that must hold across related inputs, applied successfully to ML classifiers~\cite{xie2011metamorphicml}.
We adapt this paradigm: our metamorphic relation is that \emph{renaming predicates should not change the system's ability to find a correct hypothesis}, since logical structure is preserved under bijective renaming.

\section{Background}
\label{sec:background}

\subsection{Inductive Logic Programming}

\begin{definition}[ILP Task]
An ILP task is a tuple $T = \langle B, E^+, E^-, \mathcal{H} \rangle$ where $B$ is background knowledge (a definite logic program), $E^+ = \{e_1^+, \ldots, e_m^+\}$ and $E^- = \{e_1^-, \ldots, e_n^-\}$ are sets of positive and negative examples, and $\mathcal{H}$ is the hypothesis space defined by a language bias. The goal is to find $H \in \mathcal{H}$ such that:
\begin{itemize}
    \item \textbf{Completeness}: $\forall e \in E^+: B \cup H \models e$
    \item \textbf{Consistency}: $\forall e \in E^-: B \cup H \not\models e$
\end{itemize}
\end{definition}

\subsection{Learning from Failures}

The LFF framework~\cite{cropper2021learning} decomposes the ILP task into an iterative loop:
\begin{enumerate}
    \item \textbf{Generate}: An ASP solver proposes a candidate hypothesis $H$ from $\mathcal{H}$, subject to accumulated constraints.
    \item \textbf{Test}: $H$ is evaluated against examples using Prolog. Coverage yields true positives $\text{TP}$, false positives $\text{FP}$, and false negatives $\text{FN}$.
    \item \textbf{Constrain}: Based on test results, hypotheses are pruned:
    \begin{itemize}
        \item \emph{Generalisation constraint}: if $H$ is too specific ($\text{FN} > 0, \text{FP} = 0$), prune hypotheses more specific than $H$.
        \item \emph{Specialisation constraint}: if $H$ is too general ($\text{FP} > 0$), prune hypotheses more general than $H$.
    \end{itemize}
\end{enumerate}
The loop repeats until a complete and consistent $H$ is found or $\mathcal{H}$ is exhausted.

\subsection{Hypothesis Language Bias}

The hypothesis space is defined by a \emph{bias} specification.
In Popper, this includes mode declarations \texttt{head\_pred} (the target predicate) and \texttt{body\_pred} (allowed body predicates) with argument directions as well as structural constraints such as \texttt{max\_vars}, \texttt{max\_body}, \texttt{max\_clauses}, and type signatures that tightly bound the ASP search space.

\section{The ALBI Framework}
\label{sec:method}

\begin{figure}[!ht]
\centering
\includegraphics[width=0.75\textwidth]{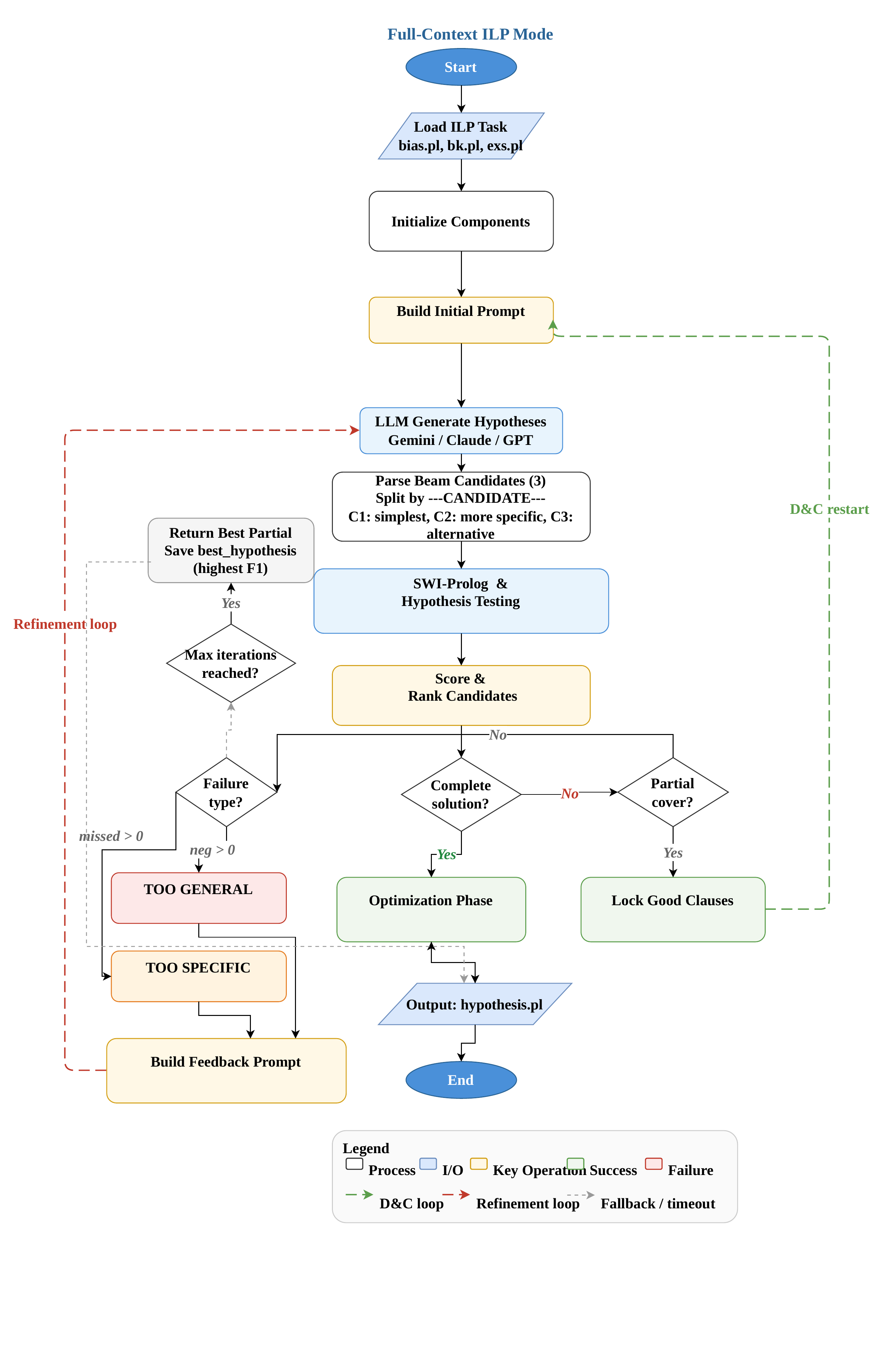}
\caption{The ALBI framework architecture.}
\label{fig:overview}
\end{figure}

ALBI replaces the ASP generator in the LFF loop with an LLM while preserving the formal test and constrain stages (Figure~\ref{fig:overview}).
The key insight is that the LLM acts as a \emph{heuristic generator} proposing candidates using semantic understanding and learned patterns while SWI-Prolog~\cite{wielemaker2012swiprolog} serves as the \emph{formal oracle}, providing rigorous coverage evaluation.
Unlike ASP-based generation, which provides completeness guarantees over $\mathcal{H}$, the LLM trades completeness for speed and semantic guidance, navigating the hypothesis space through learned inductive biases rather than exhaustive enumeration.
ALBI uses only the mode declarations from the bias, relying on the LLM's inductive priors and the iterative feedback loop rather than explicit structural bounds to guide hypothesis generation.

\subsection{Structured Prompt Architecture}
\label{sec:prompt}

The initial prompt establishes the LLM as a deterministic Prolog hypothesis generator and includes five components:

\begin{enumerate}
    \item \textbf{System role}: Instructions defining the task as ILP hypothesis induction, emphasizing deterministic Prolog syntax and the requirement for sound clauses.
    \item \textbf{Bias constraints}: The mode declarations from the task's \texttt{bias.pl}, specifying the target predicate (\texttt{head\_pred}) and allowed body predicates (\texttt{body\_pred}) with argument directions.
    \item \textbf{Background knowledge}: The complete $B$ from \texttt{bk.pl}, included in full.
    \item \textbf{Examples}: All positive ($E^+$) and negative ($E^-$) examples from \texttt{exs.pl}.
    \item \textbf{Beam request}: Generate $k = 3$ candidates ordered by specificity (simplest, more specific, alternative structure).
\end{enumerate}

Subsequent iterations use a \emph{feedback prompt} with the failed hypothesis, coverage statistics, sampled counterexamples, and accumulated constraints.

\subsection{Hypothesis Testing and Scoring}
\label{sec:testing}

Each candidate hypothesis is loaded into SWI-Prolog alongside $B$ and tested against all examples.
For each example $e$, we check whether $B \cup H \models e$ via Prolog evaluation with a timeout guard.

\begin{definition}[Scoring Function]
\label{def:score}
Given hypothesis $H$ with false positives $\text{FP}(H) = |E^- \cap \text{covered}(H)|$ and recall $r(H) = |E^+ \cap \text{covered}(H)| / |E^+|$:
\begin{equation}
\text{score}(H) = 
\begin{cases}
100 + \lfloor r(H) \times 100 \rfloor & \text{if } \text{FP}(H) = 0 \\[6pt]
\displaystyle\frac{F_1(H)}{1 + \text{FP}(H)} & \text{otherwise}
\end{cases}
\label{eq:score}
\end{equation}
where $F_1(H)$ is the harmonic mean of precision and recall over the covered examples.
\end{definition}

This scoring function prioritizes \emph{consistency} (zero false positives) over completeness, assigning scores in $[100, 200]$ for consistent hypotheses and penalizing false positives in the denominator.
Among tied candidates, we select the shortest hypothesis by clause count, implementing an MDL tie-break~\cite{rissanen1978mdl}.

\subsection{Diagnostic Feedback and Constraint Learning}
\label{sec:feedback}

When no candidate achieves a complete solution, ALBI classifies the failure (\texttt{TOO\_GENERAL}: covers negatives; \texttt{TOO\_SPECIFIC}: misses positives; \texttt{SYNTAX\_ERROR}) and generates a diagnostic prompt with the failed hypothesis, coverage statistics, sampled counterexamples (positives via exponential sampling, negatives uniform), grounded BK facts, and a new beam request.
This serves as the LFF constraint analogue: natural-language descriptions guide the LLM away from previously failed hypothesis regions.

\subsection{Locked Clauses and Partial Cover}
\label{sec:locked}

When a clause achieves \emph{partial coverage} ($\geq 1$ positive, zero false positives), ALBI \emph{locks} it: covered positives are removed from $E^+$, BK is filtered to remaining examples, and a fresh round targets the residuals.
This progressively builds multi-clause theories, analogous to Combo~\cite{cropper2023combo} but using the LLM as the combination engine.

\subsection{Divide-and-Conquer Optimization}
\label{sec:optimization}

Once a complete hypothesis is found, an optimization phase groups clauses by head predicate and prompts the LLM to produce simpler equivalents (MDL principle~\cite{grunwald2007mdl}), each verified by SWI-Prolog. Three iterations are performed; the simplest correct hypothesis is retained.
Algorithm~\ref{alg:fullcontext} summarises the full-context mode loop.

\begin{algorithm}[!ht]
\caption{ALBI Full-Context Mode}
\label{alg:fullcontext}
\begin{algorithmic}[1]
\REQUIRE Task $\langle B, E^+, E^-, \mathcal{H} \rangle$, max iterations $T$
\ENSURE Hypothesis $H^*$ or best partial solution
\STATE $\text{locked} \leftarrow \emptyset$;\; $H^* \leftarrow \bot$;\; constraints $\leftarrow \emptyset$
\STATE prompt $\leftarrow$ \textsc{BuildInitPrompt}($B, E^+, E^-, \mathcal{H}$)
\FOR{$t = 1$ \TO $T$}
    \STATE $\{h_1, h_2, h_3\} \leftarrow$ \textsc{LLM}(prompt)
    \FOR{$i = 1$ \TO $3$}
        \STATE $(\text{TP}_i, \text{FN}_i, \text{FP}_i) \leftarrow$ \textsc{TestSWI}($B, h_i, E^+, E^-$)
        \STATE $s_i \leftarrow \text{score}(h_i)$ \COMMENT{Eq.~\ref{eq:score}}
    \ENDFOR
    \STATE $h^* \leftarrow \arg\max_i s_i$
    \IF{$\text{FP}(h^*) = 0$ \AND $\text{FN}(h^*) = 0$}
        \STATE $H^* \leftarrow \text{locked} \cup h^*$
        \RETURN \textsc{Optimize}($H^*$)
    \ELSIF{$\text{FP}(h^*) = 0$ \AND $\text{FN}(h^*) \neq \emptyset$}
        \STATE locked $\leftarrow$ locked $\cup$ \textsc{GoodClauses}($h^*$)
        \STATE $E^+ \leftarrow E^+ \setminus \text{TP}(h^*)$
        \STATE $B \leftarrow$ \textsc{FilterBK}($B, E^+$)
        \STATE prompt $\leftarrow$ \textsc{BuildInitPrompt}($B, E^+, E^-, \mathcal{H}$, constraints)
    \ELSE
        \STATE constraints $\leftarrow$ constraints $\cup$ \textsc{AnalyseFailure}($h^*$)
        \STATE prompt $\leftarrow$ \textsc{BuildFeedback}($h^*, \text{TP}, \text{FN}, \text{FP}$)
    \ENDIF
\ENDFOR
\RETURN locked $\cup$ $H^*_{\text{best}}$
\end{algorithmic}
\end{algorithm}

\subsection{Semantic Anonymization}
\label{sec:anonymization}

To rigorously test whether ALBI performs structural symbolic induction rather than exploiting semantic cues in predicate names (e.g., \texttt{parent(X,Y)} suggests a family relationship), we introduce \emph{semantic anonymization}:

\begin{definition}[Semantic Anonymization]
Given an ILP task $T = \langle B, E^+, E^-, \mathcal{H} \rangle$ with predicate vocabulary $\mathcal{P}$, the anonymized task $T' = \langle B', E'^+, E'^-, \mathcal{H}' \rangle$ is obtained by applying a bijective mapping $\sigma: \mathcal{P} \to \{p_1, p_2, \ldots, p_{|\mathcal{P}|}\}$ to all predicate names in $B$, $E^+$, $E^-$, and the bias specification.
\end{definition}

The metamorphic relation~\cite{chen2018metamorphic} is: if the system correctly solves $T$, then it should also correctly solve $T'$, since the logical structure is invariant under bijective predicate renaming.
A system whose success rate drops significantly on $T'$ relative to $T$ relies on surface-level semantic pattern matching rather than structural inductive reasoning.
We treat semantic anonymization as an ablation of predicate-name semantics: each task is evaluated twice, once with original names and once with anonymized identifiers. This preserves the logical structure while removing surface semantic cues, allowing us to measure how much ALBI depends on predicate-name meaning.

\section{Experiments}
\label{sec:experiments}

\subsection{Experimental Setup}
\label{sec:setup}

\paragraph{Benchmarks.}
We evaluate on \textbf{30} standard ILP tasks from the Popper benchmark suite\footnote{\url{https://github.com/logic-and-learning-lab/Popper}}~\cite{cropper2021learning}, spanning five domains: program synthesis (10 tasks), IGGP game playing (9), Zendo spatial reasoning (3), robots (3), and other (5).
All tasks use noise-free examples and the standard Popper format (\texttt{bias.pl}, \texttt{bk.pl}, \texttt{exs.pl}).
Each task has both a \emph{semantic} variant (original predicate names) and an \emph{anonymized} variant (abstract identifiers $p_0, p_1, \ldots$).
ALBI extracts only mode declarations from \texttt{bias.pl}; Popper's additional constraints (\texttt{max\_vars}, \texttt{max\_body}, \texttt{max\_clauses}, type restrictions) are not used, making ALBI's bias strictly weaker.

\paragraph{Models.}
We evaluate three frontier LLMs (Table~\ref{tab:models}): \textbf{Gemini~3.1~Pro} ($\tau{=}0$, thinking level high), \textbf{Claude~Opus~4.6} ($\tau{=}1.0$, extended thinking with 112K budget), and \textbf{GPT~5.4~Pro} (reasoning effort ``high'', 48K output limit). All use structured XML prompts with $k{=}3$ beam candidates and failure-specific feedback.

\begin{table}[t]
\centering
\caption{LLM API configurations.}
\label{tab:models}
{\small
\begin{tabular}{@{}lcll@{}}
\toprule
\textbf{Model} & $\tau$ & \textbf{Reasoning Mode} & \textbf{Max Out} \\
\midrule
Gemini 3.1 Pro    & 0.0 & Thinking: high         & Default \\
Claude Opus 4.6   & 1.0 & Ext.\ thinking (112K)  & 128K \\
GPT 5.4 Pro       & --  & Effort: high           & 48K \\
\bottomrule
\end{tabular}
}
\end{table}

\paragraph{Baseline.}
We compare against \textbf{Popper}~\cite{cropper2021learning} (Clingo-based ASP generation, 1200\,s timeout per task). ALBI terminates when a correct hypothesis is found or the maximum iteration count $T{=}20$ is reached; beam width $k{=}3$ balances diversity against API cost (higher $k$ increases latency with diminishing returns on this benchmark).

\paragraph{Metrics.}
(i)~\textbf{Success rate}: fraction of tasks yielding a complete and consistent hypothesis.
(ii)~\textbf{Iterations}: number of generate-test cycles.
(iii)~\textbf{Wall-clock time}: total elapsed time including API latency.
(iv)~\textbf{Hypothesis size}: clause count and total literal count (MDL proxy).

\subsection{Main Results}
\label{sec:main_results}

\begin{table}[t]
\centering
\caption{Overall performance across 30 semantic benchmark tasks.}
\label{tab:main}
{\small
\begin{tabular}{@{}lrrrr@{}}
\toprule
\textbf{System} & \textbf{Succ.~(\%)} & \textbf{Iter.} & \textbf{Time~(s)} & \textbf{Clauses} \\
\midrule
Popper (ASP)      & 100.0 & 6.8 & 81.8   & 2.7 \\
ALBI + Gemini 3.1 Pro    & 100.0 & 1.7 & 333.6  & 2.8 \\
ALBI + GPT 5.4 Pro & 100.0 & 1.4 & 1631.9 & 2.8 \\
ALBI + Claude Opus 4.6    & 100.0 & 1.5 & 397.7  & 2.9 \\
\bottomrule
\end{tabular}
}
\end{table}

Table~\ref{tab:main} summarises the main results.
All four systems achieve 100\% success across all five domains (program synthesis, IGGP, Zendo, robots, other), solving all 30 tasks.
Three different frontier LLMs, each using the same ALBI framework, match the completeness of Popper's exhaustive ASP search on this benchmark suite.

\paragraph{Iteration efficiency.}
The most striking result is ALBI's iteration count.
All three LLM configurations converge in 1.4--1.7 iterations on average, meaning the vast majority of tasks are solved on the \emph{first attempt} without feedback.
This confirms that frontier LLMs possess strong inductive biases for logic program synthesis when provided with structured prompts containing background knowledge, examples, and bias constraints.
Popper, by contrast, iterates through a combinatorial ASP search (mean 6.8 refinement steps across solved tasks), though its per-iteration cost is far lower.

\paragraph{Time--accuracy trade-off.}
Popper is the fastest system (81.8\,s mean), as its search involves no network latency.
Among ALBI configurations, Gemini is fastest (333.6\,s), followed by Claude (397.7\,s), with GPT~5.4~Pro significantly slower (1631.9\,s) due to higher API latency and longer generation times.
Notably, GPT~5.4~Pro's higher latency does not reduce its accuracy it achieves 100\% success suggesting that the additional computation may reflect more thorough reasoning.

\paragraph{Hypothesis compactness.}
All systems produce comparably sized hypotheses (2.7--2.9 clauses on average).
However, GPT~5.4~Pro achieves the lowest mean literal count (6.0 vs.\ 8.2--8.6 for Claude and Gemini), and its optimization phase successfully reduces literals in 12/12 applicable tasks.
This suggests GPT~5.4~Pro's structured reasoning favours more concise logical representations.

\begin{figure}[t]
\centering
\includegraphics[height=3.8cm]{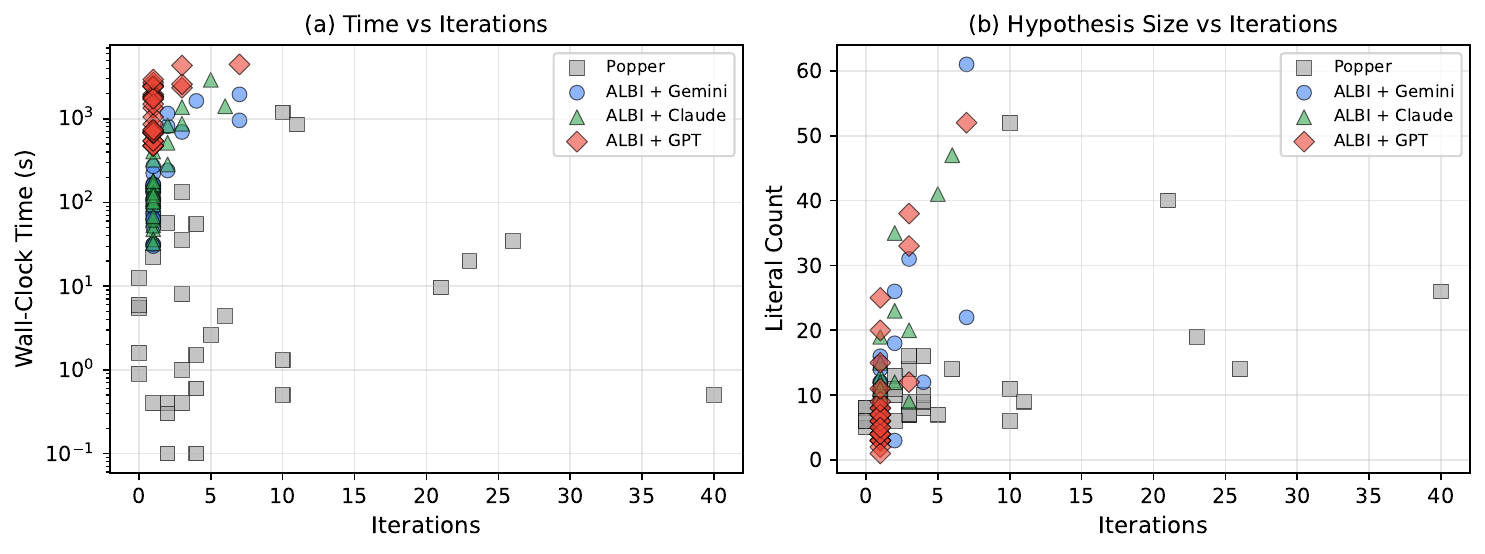}
\caption{Wall-clock time and literal count vs.\ iteration count per system.}
\label{fig:faceted}
\end{figure}

The cost structure differs fundamentally between systems: ALBI models cluster at 1--2 iterations with 40--3000\,s per task (dominated by API latency), while Popper spreads across up to 40 cheap ASP iterations.
ALBI's per-iteration cost is 10--100 times higher, but it requires 5--20 times fewer iterations.
Hypothesis complexity (literal count) is largely independent of iteration count, suggesting that iteration count reflects search difficulty rather than target concept complexity (Figure~\ref{fig:faceted}).

\subsection{Optimization Effectiveness}
\label{sec:optimization_results}

ALBI's divide-and-conquer optimization phase (Section~\ref{sec:optimization}) attempts to simplify the initial hypothesis while preserving coverage.
Table~\ref{tab:optimization} reports the optimization rates across models.

\begin{table}[t]
\centering
\caption{Optimization effectiveness across models.}
\label{tab:optimization}
{\small
\begin{tabular}{@{}lrrrr@{}}
\toprule
\textbf{System} & \textbf{Applicable} & \textbf{Optimized} & \textbf{Rate (\%)} & \textbf{Literals Reduced} \\
\midrule
ALBI + Gemini      & 15/30 & 4/30  & 13.3  & $-$26 \\
ALBI + Claude      & 15/30 & 1/30  & 3.3   & $-$7 \\
ALBI + GPT 5.4 Pro & 12/30 & 12/30 & 40.0  & $-$87 \\
\bottomrule
\end{tabular}
}
\end{table}

Of the 30 tasks, only a subset triggers the D\&C optimization phase (tasks where the initial hypothesis has multi-clause structure amenable to simplification); the remaining tasks are solved directly without optimization.
GPT~5.4~Pro is uniquely effective: all 12 applicable tasks are successfully simplified (mean reduction of 7.3 literals per task).
Inspection of the hypotheses reveals why: during optimization, GPT systematically rewrites list-manipulating clauses using Prolog's native pattern matching (\texttt{[H|T]}) instead of explicit \texttt{head/2} and \texttt{tail/2} predicates, e.g.\ reducing \texttt{f(A)\,:- head(A,B), even(B), tail(A,C), f(C)} to \texttt{f([A|B])\,:- even(A), f(B)}.
Across the 10 program synthesis tasks, GPT uses list notation in 9/10 (semantic) and 8/10 (anonymized); Gemini uses it in only 1/10, and Claude never adopts it.
This notation is unavailable to Popper's ASP generator and is not declared in the bias, making it a concrete instance of LLMs extending the effective hypothesis language beyond the formal specification.
Gemini shows moderate optimization success (4 of 15 applicable), while Claude rarely improves its initial hypotheses (1 of 15), suggesting that Claude's first-pass solutions are already near-minimal.

\subsection{Semantic Anonymization Results}
\label{sec:anon_results}

\begin{table}[t]
\centering
\caption{Effect of semantic anonymization on success rate (\%) and average iterations.}
\label{tab:anon}
{\small
\begin{tabular}{@{}lrrrr@{}}
\toprule
 & \multicolumn{2}{c}{\textbf{Semantic}} & \multicolumn{2}{c}{\textbf{Anonymized}} \\
\cmidrule(lr){2-3} \cmidrule(lr){4-5}
\textbf{System} & Succ.~(\%) & Iter. & Succ.~(\%) & Iter. \\
\midrule
ALBI + Gemini      & 100.0 & 1.7 & 100.0 & 2.6 \\
ALBI + GPT 5.4 Pro & 100.0 & 1.4 & 100.0 & 1.9 \\
ALBI + Claude      & 100.0 & 1.5 & 100.0 & 2.5 \\
\bottomrule
\end{tabular}
}
\end{table}

Table~\ref{tab:anon} reports the impact of the semantic-anonymization ablation, in which each task is run twice: once with meaningful predicate names and once with anonymized identifiers (Popper excluded its ASP search is invariant to naming).
All three models maintain \textbf{100\% success}, providing evidence of structural induction rather than reliance on predicate-name semantics.
The iteration multiplier quantifies increased difficulty: Gemini 1.53 times (1.7$\to$2.6), Claude 1.67 times (1.5$\to$2.5), GPT~5.4~Pro 1.36 times (1.4$\to$1.9).
GPT~5.4~Pro is the most robust, suggesting its reasoning relies more on structural patterns than surface semantics.

\begin{figure}[t]
\centering
\includegraphics[width=\textwidth]{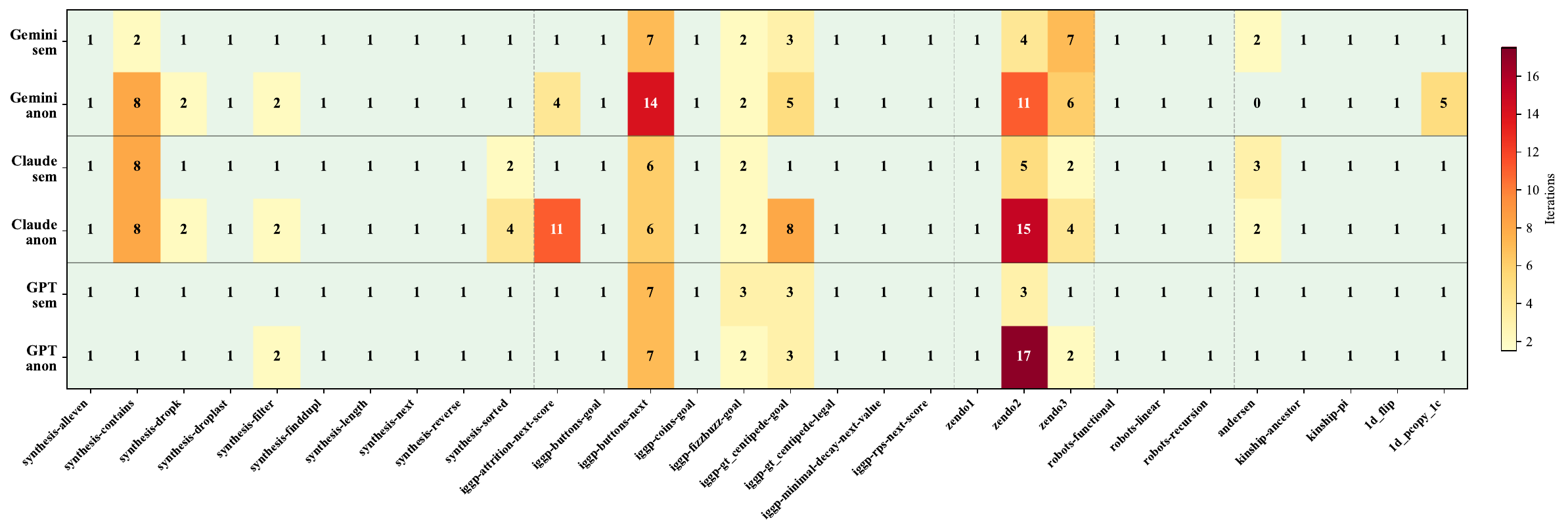}
\caption{Per-task iteration counts (semantic vs.\ anonymized). Tasks on X-axis; model$\times$condition on Y-axis. Green = single-iteration; darker = harder. Dashed lines separate domains.}
\label{fig:iter_heatmap}
\end{figure}

Figure~\ref{fig:iter_heatmap} shows that most tasks are solved in a single iteration; the hardest (\texttt{zendo2}, \texttt{iggp-buttons-next}, \texttt{synthesis-contains}) require substantially more iterations under anonymization.

\FloatBarrier

\subsection{Qualitative Analysis}
\label{sec:qualitative}

We highlight four illustrative cases.

\paragraph{Case 1: Semantic bias and optimization risk (\texttt{andersen}).}
Gemini produces the textbook-correct four-rule solution for Andersen's points-to analysis in one iteration, including the recursive store rule that Popper cannot distinguish from over-general alternatives (zero negative examples). Popper's solution exploits coincidental BK patterns and would fail on larger programs.
However, GPT~5.4~Pro's D\&C optimization incorrectly simplifies a correct rule to pass the test set, revealing that literal-count minimization can damage hypotheses when negatives are insufficient.

\paragraph{Case 2: Feedback-driven recovery (\texttt{1d\_flip}).}
Popper requires 1186\,s and 10 refinements; all ALBI models converge in 1--2 iterations, demonstrating that LLM heuristics bypass combinatorial explosion on structurally complex tasks.

\paragraph{Case 3: Constant reasoning under anonymization (\texttt{iggp-attrition-next-score}).}
With semantic names, Gemini solves this in one iteration (3~rules, 8~literals) by reasoning about action constants. Under anonymization, 4~iterations and 18~literals are needed showing that constant semantics accelerate convergence but are not required for correctness.

\paragraph{Case 4: Built-in primitives (\texttt{zendo2}).}
ALBI uses inequality (\texttt{\textbackslash=}) not declared in the bias to produce a 2-rule, 12-literal solution in 4~iterations; Popper requires 26~iterations without access to this primitive. Under anonymization, 11~iterations and 7~rules are needed, illustrating that LLM knowledge of Prolog idioms acts as an implicit bias extension.

\section{Discussion}
\label{sec:discussion}

\paragraph{ALBI vs.\ Popper.}
ALBI's near-first-shot convergence (1.4--1.7 iterations) demonstrates that LLMs can shortcut the combinatorial search that ASP solvers traverse exhaustively.
Crucially, ALBI achieves this with a \emph{strictly weaker bias}: only mode declarations are provided, whereas Popper additionally relies on structural constraints (\texttt{max\_vars}, \texttt{max\_body}, \texttt{max\_clauses}) and type restrictions.
The LLM's learned inductive priors, combined with iterative feedback, substitute for hand-crafted hypothesis space bounds.
Additionally, the LLM can employ built-in Prolog primitives (inequality, negation-as-failure) not declared in the bias, effectively extending the hypothesis language (cf.\ Case~4 in Section~\ref{sec:qualitative}).
However, Popper provides completeness guarantees and is 3--20 times faster in wall-clock time; for production settings where latency and cost matter, Popper remains practical for tasks within its search capacity.

\paragraph{Comparison fairness and the semantic gap.}
LLMs are pretrained on data that may include ILP-related content, giving ALBI an implicit advantage over Popper's purely algorithmic search.
We view this as a \emph{feature} of the neuro-symbolic integration: leveraging LLM semantic heuristics while retaining formal verification is the core motivation.
Our anonymization experiments control for the strongest form of this concern: predicate names accelerate convergence by 1.4--1.75 times but 100\% success is maintained without them, demonstrating genuine structural induction rather than surface-level pattern matching.
The iterative feedback mechanism is the key enabler even when the LLM's initial hypothesis is wrong on anonymized tasks, structured diagnostic feedback (coverage statistics, counterexamples) allows recovery within 2--3 iterations.

\paragraph{Difficulty analysis.}
Iteration counts reveal meaningful difficulty gradients under anonymization.
The hardest tasks are \texttt{zendo2} (11--17 iterations vs.\ 1--7 semantic), \texttt{iggp-buttons-next} (up to 14 vs.\ 7), and \texttt{synthesis-contains} (8 for Gemini vs.\ 2).
No task exceeds the 20-iteration budget, but the 100\% ceiling is partly an artifact of evaluating only on Popper's own benchmark suite.
The iteration data also provides implicit component analysis: under semantic conditions 40\% of tasks require $>$1 iteration; under anonymization this rises to 57\% (essential feedback loop). Multi-clause tasks succeed via the locked clause mechanism, and anonymization constitutes a controlled ablation of semantic priors (iterations increase 1.36--1.67 times, success preserved). No task exhausts the $T{=}20$ iteration budget, confirming that the feedback mechanism is sufficient for recovery on all 30 tasks.

\paragraph{Limitations.}
Key limitations include: (i)~no completeness guarantee the LLM may fail on structures it has not learned; (ii)~limited reproducibility across API versions; (iii)~3--20 times higher wall-clock cost than Popper due to API latency; (iv)~optimization risk D\&C can damage correct hypotheses with insufficient negatives; (v)~noise-free assumption integrating MDL-based scoring~\cite{hocquette2024noisy} is planned; (vi)~the weaker bias trades optimality guarantees for richer hypothesis spaces, and each LLM call is orders of magnitude more expensive than an ASP iteration.

\section{Conclusion}
\label{sec:conclusion}

We presented ALBI, a neuro-symbolic ILP framework that integrates Large Language Models into the Learning from Failures paradigm.
ALBI replaces the ASP generator with an LLM-guided generate-test-constrain loop, retaining SWI-Prolog as a formal verifier to combine the semantic heuristics of neural models with the logical rigor of symbolic reasoning.
Experiments on 30 standard ILP benchmark tasks demonstrate that ALBI matches Popper's 100\% success rate across all three LLMs while converging in only 1.4--1.7 iterations on average.
Our semantic anonymization methodology provides evidence that ALBI performs robust symbolic induction: all models maintain 100\% success under anonymization, with only a moderate increase in iterations (1.36--1.67 times).
\paragraph{Future Work.}
Key extensions include: batch-based scalability for large tasks, LLM-guided predicate invention, noisy-example support via MDL selection, per-model hyperparameter optimization, a cross-model \emph{Logic Robustness Index}, and evaluation on broader ILP datasets beyond the Popper suite.

\paragraph{Appendix: Prompt Template Summary.}
The initial prompt ($t{=}1$) contains six components:
\emph{(i)~System role} ``You are an expert ILP solver; synthesize a Prolog hypothesis that entails all positives, rejects all negatives, and follows Occam's Razor'';
\emph{(ii)~Strategic guidelines} identify the target predicate, enforce variable sharing, generalize constants, choose recursion vs.\ disjunction;
\emph{(iii)~Bias} mode declarations from \texttt{bias.pl} (\texttt{head\_pred}/\texttt{body\_pred} with typed directions);
\emph{(iv)~Background knowledge} complete $B$ from \texttt{bk.pl}, included verbatim;
\emph{(v)~Examples} all $E^+$/$E^-$ from \texttt{exs.pl};
\emph{(vi)~Beam request} generate $k{=}3$ candidates (simplest, more specific, alternative), each delimited by \texttt{---CANDIDATE---}.
The feedback prompt ($t>1$) replaces (iv)--(v) with:
the previous best candidate with coverage statistics ($\text{TP}$, $\text{FP}$, $\text{FN}$);
a failure classification (\texttt{TOO\_GENERAL}, \texttt{TOO\_SPECIFIC}, or \texttt{SYNTAX\_ERROR});
sampled counterexamples with grounded BK facts;
analysis hints (compare BK for positives vs.\ negatives, find discriminating literals); and accumulated constraint history.
The system prompt instructs the LLM to produce a one-sentence pattern analysis before the hypothesis, encouraging chain-of-thought reasoning. Common-mistake warnings (e.g., anonymous head variables, disconnected body variables) are included to prevent frequent errors.

\FloatBarrier
\bibliographystyle{splncs04}
\bibliography{IJCLR2026_Paper}

\end{document}